\documentclass[letterpaper]{article} % DO NOT CHANGE THIS
\usepackage{aaai2027}  % Final, non-anonymous version
\nocopyright
\usepackage[hyphens]{url}  % DO NOT CHANGE THIS
\usepackage{graphicx} % DO NOT CHANGE THIS
\usepackage{natbib}  % DO NOT CHANGE THIS AND DO NOT ADD ANY OPTIONS TO IT
\usepackage{caption} % DO NOT CHANGE THIS AND DO NOT ADD ANY OPTIONS TO IT
\usepackage{algorithm}
\usepackage{algorithmic}

\usepackage{newfloat}
\usepackage{amsmath}
\usepackage{listings}
\DeclareCaptionStyle{ruled}{labelfont=normalfont,labelsep=colon,strut=off} % DO NOT CHANGE THIS
\floatstyle{ruled}
\newfloat{listing}{tb}{lst}{}
\floatname{listing}{Listing}

\usepackage{booktabs}

\title{Complementary Retrieval-Augmented Prompting for Consistent Long-Form Video Generation}
\author{
    Xianghan Wei, Xiaoda Yang, Zhi Wang, An Pan, Daoan Zhang,\\
    Huayi Zhang, Yan Zhang, Wei Xu, Zishun Liao, Jianwen Lou
}
\affiliations{
}

\begin{document}

\maketitle

\begin{abstract}
While recent video foundation models excel at generating high-quality short videos, long-form video generation remains challenging because independently generated shots must preserve consistent characters, scenes, and objects throughout a story. Existing training-free approaches typically condition target shots using retrieved historical visuals. However, these references often suffer from severe informational mismatch, either introducing irrelevant contextual redundancy or failing to provide the complete set of elements required for the target shot. To resolve this, we present Complementary Retrieval-Augmented Prompting, an agentic framework that strategically aggregates a compact set of mutually supportive historical references to achieve complete and targeted conditioning for long-form video generation without retraining or modifying the underlying generator. Specifically, our framework explicitly models the visual elements required by each target shot by parsing the narrative script into a text-grounded visual element registry that tracks characters, objects, scenes, and their shot-level states. A VLM-annotated keyframe library further maps these elements to past visual observations. Guided by the required elements, our agent retrieves complementary references that maximize target-element coverage while minimizing historical noise. Finally, the retrieved references, structured element states, and grounding instructions are assembled into a unified prompt for the frozen video generator. This element-aware process provides comprehensive conditioning while remaining fully interpretable. Quantitative and qualitative evaluations on multi-shot story generation demonstrate improved cross-shot character, object, and scene consistency over memory-based and agentic retrieval baselines.

\end{abstract}


\section{Introduction}
Recent video foundation models can synthesize visually rich short clips from text, images, videos, and other multimodal prompts~\citep{gao2025seedance,klingteam2025klingomni,huang2025stepvideo,bao2024vidu,alibaba2025wan26,googledeepmind2025veo31}. Rather than training a specialized long-video model, creators can decompose a story into shots, generate each shot with an off-the-shelf reference-conditioned model, and assemble the resulting clips into a complete narrative. This shot-by-shot workflow is already widely used in the production of short dramas and motion comics because it is flexible, interpretable, and compatible with rapidly evolving commercial video APIs.

However, long-form generation is not simply a matter of stitching short clips together. When each shot is generated independently, recurring characters may drift in appearance, objects may disappear or change shape, and scenes may be reintroduced with inconsistent layouts~\citep{zhang2025storymem,he2026entitybench}. Reference images and videos can mitigate these issues, but they also introduce a new bottleneck: before generating each shot, the system must decide which historical references are useful, which visual elements in those references should be preserved, and which stale elements should be suppressed. In practice, this reference selection and prompt construction process is still largely manual, especially when multiple characters, objects, and locations recur over long temporal gaps.

Existing long-form video methods address consistency from different directions, but they leave this reference-management problem underexplored for frozen multimodal video generators. Training-based approaches propagate memory through attention, caches, latent retrieval, or learned conditioning~\citep{meng2025holocine,wu2025cinetrans,an2025onestory,luo2026shotstream,zhang2025storymem,wei2026memento,hu2026longliverag}, which often incur high computational overhead and lack scalability. Agentic and explicit-memory pipelines improve controllability by maintaining scripts, entities, or generated keyframes~\citep{lin2023videodirectorgpt,zheng2024vgot,wu2025movieagent,huang2026vimax,liu2026reca,zhou2026videomemory,yin2026cotrisygen,lai2026groundshot}, but they often rely on predefined full-script analysis, entity-level canonical references, or manually designed prompt/reference layouts. These designs are powerful, yet less suited to incremental and interactive creation where users may extend or revise the story while generation proceeds.

We propose \emph{Complementary Retrieval-Augmented Prompting}, an agentic framework that treats long-form video generation as an interpretable retrieval and multimodal prompting problem. The agent maintains two evolving evidence structures. First, a text-grounded visual element registry tracks characters, objects, and scenes introduced by the script, together with shot-level states such as should-reference, should-exclude, optional, and newly introduced. Second, a VLM-annotated keyframe library records where these elements appear in generated shots, using both element-level annotations and holistic frame descriptions. Given a new shot, the agent retrieves a compact set of complementary historical keyframes that jointly cover the required visual elements while penalizing references that may introduce conflicting content. These selected references are then converted into a structured multimodal prompt that explicitly explains how each image should be used or avoided by the frozen generator.

This formulation has two practical advantages. First, it makes reference usage explainable: each selected frame can be traced to the visual elements it covers and the risks it may introduce. Second, it supports online and interactive generation: the registry and keyframe library are updated after each generated shot, so the system does not require a complete script or a fixed generation schedule in advance. Long-range visual consistency is handled by retrieval over historical evidence, while local temporal continuity between non-cut shots can be handled separately through proximal video references and lightweight boundary smoothing.

Our contributions are threefold:
\begin{itemize}
\item We introduce a training-free, online framework that maintains a visual element registry and a VLM-annotated keyframe library, turning generated history into structured and editable evidence without requiring the full script in advance.

\item We formulate reference selection as a complementary coverage problem and develop a greedy retrieval algorithm that selects compact historical references covering target elements while suppressing conflicting content.

\item We ground each retrieved frame through element-level and holistic guidance in a structured multimodal prompt. Experiments on EntityBench demonstrate improved cross-shot consistency over training-based methods and explicit workflow baselines.
\end{itemize}

\section{Related Work}

\subsection{Training-based Long-Form Video Generation}

Training-based methods embed cross-shot consistency into the generator itself. HoloCine~\citep{meng2025holocine} and CineTrans~\citep{wu2025cinetrans} jointly generate multiple shots and leverage cross-shot attention to preserve visual consistency, while autoregressive and memory-based approaches propagate selected frames, feature caches, attention sinks, reconstructed subjects, or retrieved latents across generation steps~\citep{an2025onestory,luo2026shotstream,zhang2025storymem,wei2026memento,chen2026longlive2,hu2026longliverag}. Although effective, these mechanisms require model-specific training or access to internal representations, limiting both their portability across video generators and the interpretability of their historical conditioning.

\subsection{Explicit and Agentic Long-Form Generation}

Agentic pipelines organize existing generators through script decomposition, shot planning, context allocation, and iterative visual feedback~\citep{lin2023videodirectorgpt,zheng2024vgot,wu2025movieagent,huang2026vimax,liu2026reca}. Entity-centric systems further maintain explicit states for recurring characters, objects, and scenes~\citep{zhou2026videomemory,yin2026cotrisygen}. Meanwhile, reference-conditioned generators such as Seedance, Kling-Omni, Step-Video, and Vidu support multimodal prompting over text, images, and videos~\citep{gao2025seedance,klingteam2025klingomni,huang2025stepvideo,bao2024vidu}, but long-form workflows still rely heavily on manual reference selection and prompting. GroundShot~\citep{lai2026groundshot} reduces this burden by analyzing the full script, scheduling entity-reference shots first, and grounding subsequent shots on the resulting entity memory, but still relies on a predefined complete generation task.

Our framework instead supports incremental creation without a complete script or predefined generation order. It treats generated keyframes as VLM-annotated evidence, retrieves complementary frames under a limited reference budget, and explicitly grounds how each frame should be used while suppressing obsolete or conflicting content. This yields an interpretable and model-agnostic interface for frozen multimodal video generators.

\begin{figure*}[t]
\centering
\includegraphics[width=\textwidth]{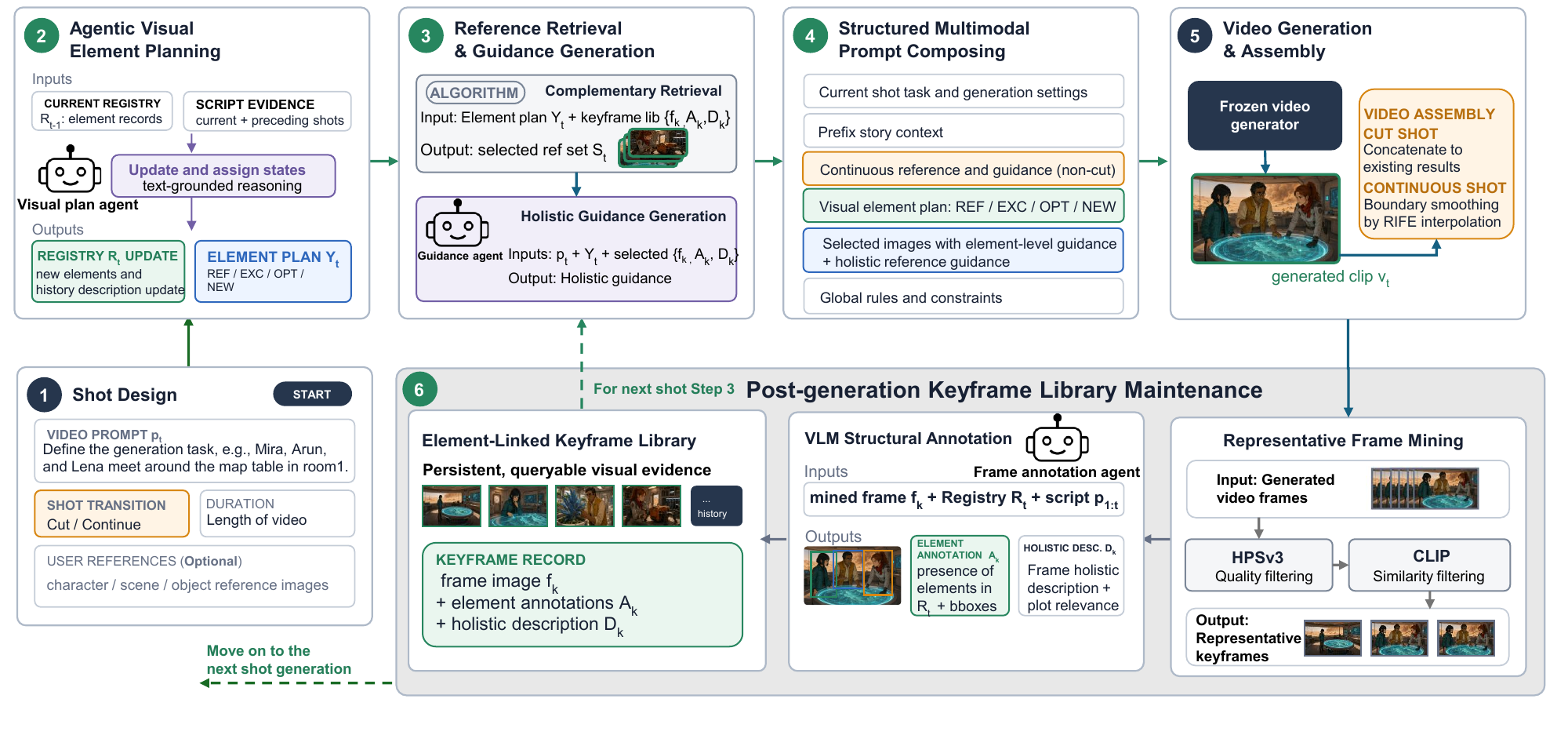}
\caption{Overview of Complementary Retrieval-Augmented Prompting. An LLM updates the visual element registry and shot-level plan; greedy coverage retrieves complementary keyframes from the VLM-annotated library; and a second LLM produces per-reference guidance for structured prompting. After generation, representative frames are mined and annotated before the library is updated. The dashed arrow denotes retrieval from the updated library for the next shot.}
\label{fig:pipeline}
\end{figure*}

\section{Method}

\subsection{Overview}
\label{subsec:overview}
We study long-form story video generation by leveraging the exceptional visual quality and robust physical modeling capabilities of frozen short-video generators. Given a narrative script decomposed into a sequence of $T$ shots $\mathcal{S} = \{s_1, \ldots, s_T\}$, each shot $s_t$ is generated through an independent inference call, and the generator retains no internal memory across shots. The central challenge is how to strategically construct a targeted multimodal prompt with historical references to preserve the consistency of characters, objects, and scenes across shots. To address this challenge, we present \textit{Complementary Retrieval-Augmented Prompting}, an agentic framework designed to strategically orchestrate inference-stage historical conditioning. Instead of relying on passive, recency-biased retrieval, our method formalizes reference selection as an optimization problem where historical frames are selected if and only if they \textit{collectively} maximize the coverage of the target shot's required elements while minimizing informational redundancy. Under this formulation, an agent continuously tracks structured visual evidence and retrieves a compact set of \textit{complementary} historical references to drive the generation loop. As illustrated in Figure~\ref{fig:pipeline}, our framework consists of four sequential modules:
\begin{enumerate}
    \item \textbf{Narrative-Grounded Element Registry:} Parses the narrative script into a structured registry tracking required elements and their shot-level states.
    \item \textbf{Element-Linked Keyframe Library:} Employs a VLM to annotate and index representative historical keyframes, establishing explicit cross-modal mappings for a high-diversity candidate reference pool.
    \item \textbf{Element-Aware Complementary Retrieval:} Executes a strategic agentic search to select a minimal, mutually supportive subset of historical references that covers target elements.
    \item \textbf{Unified Prompt Assembly:} Compiles the retrieved references, structured states, and grounding instructions into a cohesive prompt for the frozen generator.
\end{enumerate}

\paragraph{Shot Specification.} We represent each shot $s_t$ as a four-field configuration tuple:
\begin{equation}
\label{eq:shot_spec}
s_t = (p_t, d_t, c_t, r_t),
\end{equation}
where $p_t$ is the foundational text prompt directing scenes, characters, and actions, and $d_t$ denotes the target clip duration. The binary indicator $c_t \in \{\text{true}, \text{false}\}$ specifies the relationship to the preceding shot: $c_t = \text{true}$ indicates that the current shot does not continue from the preceding shot, whereas $c_t = \text{false}$ indicates that it continues from the end of the preceding shot. Finally, $r_t$ represents optional external reference assets (e.g., initial character sheets) used to anchor identity; it is not used in our algorithm or experiments.

\subsection{Narrative-Grounded Element Registry}
\label{subsec:registry}
To transform an unstructured narrative script into explicitly computable constraints for consistent long-form synthesis, our framework maintains an \textit{Element Registry} $\mathcal{R} = \{e_i\}_{i=1}^N$. Instead of relying on implicit, text-driven prompt adherence, which often suffers from text-to-video drift, this registry operationalizes reusable storytelling elements in a centralized, structured metadata store. Each element $e_i \in \mathcal{R}$ is defined as a tuple:
\begin{equation}
\label{eq:element_def}
e_i = (\text{id}_i, \text{name}_i, \text{type}_i, t_{\text{intro}, i}, \text{attr}_i),
\end{equation}
where $\text{id}_i$ is a stable unique identifier, $\text{name}_i$ provides a concrete textual definition (e.g., ``the blond boy''), $\text{type}_i \in \{\text{scene}, \text{character}, \text{object}\}$, $t_{\text{intro}, i}$ records the shot index of its first introduction, and $\text{attr}_i$ stores its persistent visual attributes.

\paragraph{Agentic Element Tracking.} The LLM agent continuously tracks element metadata and assigns conditional states $y_{i,t} \in \mathcal{Y} = \{\texttt{REF}, \texttt{EXC}, \texttt{OPT}, \texttt{NEW}\}$ according to the target shot prompt $p_t$ through a two-step procedure:
\begin{enumerate}
    \item \textbf{Registry Extension:} The agent parses $p_t$ to detect novel storytelling elements. Upon identification, a new element $e_i$ is instantiated with $t_{\text{intro}, i} = t$, assigned $y_{i,t}=\texttt{NEW}$, and appended to $\mathcal{R}$.
    \item \textbf{State Assignment:} For all pre-existing elements $\{e_i \in \mathcal{R} \mid t_{\text{intro}, i} < t\}$, the agent evaluates the target narrative context and assigns a shot-level constraint state:
    \begin{itemize}
        \item $y_{i, t} = \texttt{REF}$ (\textit{should-reference}): Explicit evidence indicates the element must appear in shot $s_t$, instructing downstream modules to maximize its historical visual coverage.
        \item $y_{i, t} = \texttt{EXC}$ (\textit{should-exclude}): Explicit evidence indicates that the element should be absent from shot $s_t$, instructing downstream modules to penalize and suppress historical visual leakage.
        \item $y_{i, t} = \texttt{OPT}$ (\textit{optional-or-uncertain}): The narrative provides no definitive evidence for presence or absence, permitting flexible contextual adaptation.
    \end{itemize}
\end{enumerate}

\begin{figure}[t]
\centering
\includegraphics[width=\columnwidth]{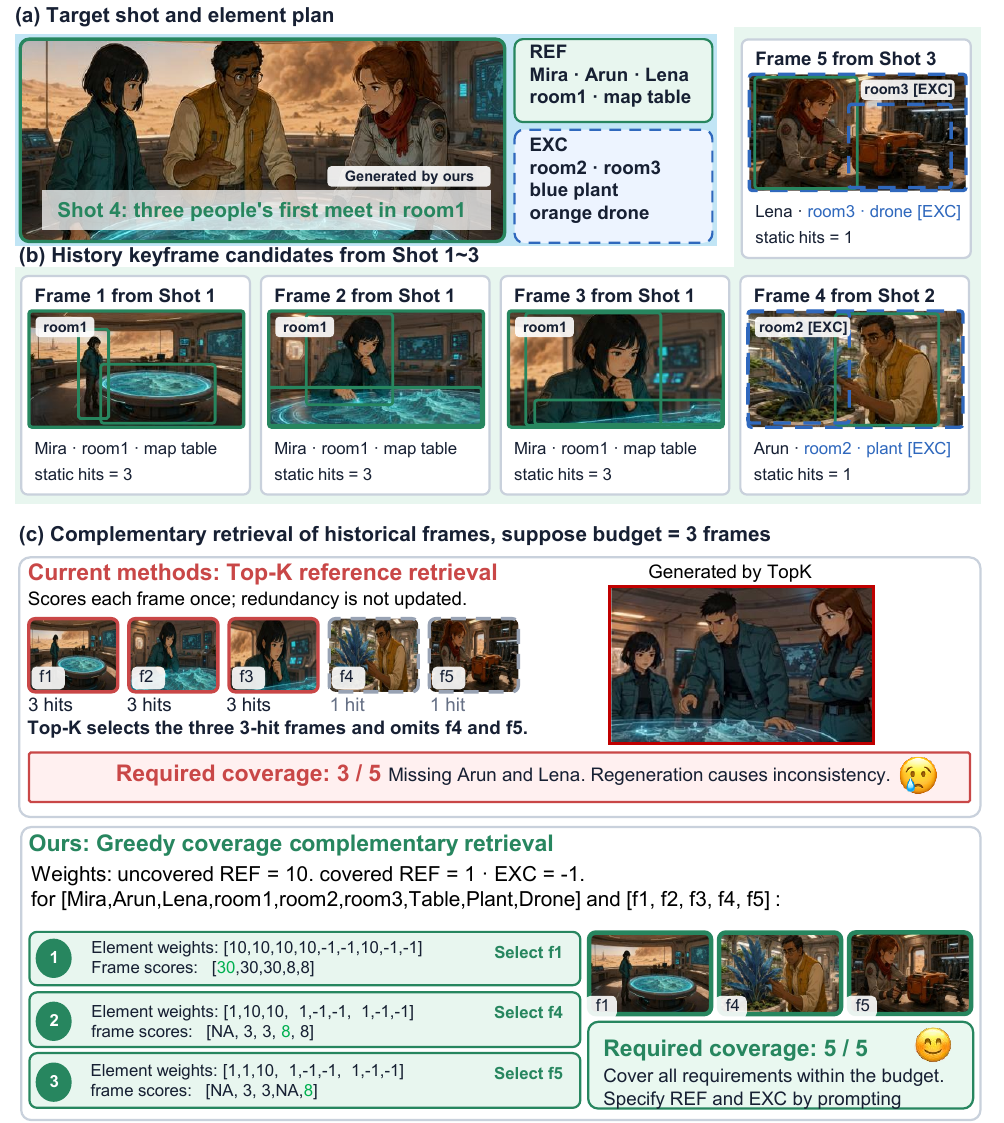}
\caption{Complementary reference retrieval example. Top-$K$ spends its three-frame budget on redundant views $f_1$--$f_3$ and covers only three of five required elements. Iterative reweighting instead selects $f_1$, $f_4$, and $f_5$, covering all required elements.}
\label{fig:complementary_example}
\end{figure}

Newly introduced elements ($t_{\text{intro}, i} = t$) are marked as \texttt{NEW} and bypass retrieval because they have no prior visual observations. As illustrated in Figure~\ref{fig:complementary_example}(a), the resulting state vector $\mathbf{Y}_t = \{y_{i, t}\}_{i=1}^{|\mathcal{R}|}$ transforms the unstructured textual prompt $p_t$ into an element-aware query representation that guides downstream complementary retrieval.

\subsection{Element-Linked Keyframe Library}
\label{subsec:library}
To establish a structured and queryable visual memory, our framework constructs an \textit{Element-Linked Keyframe Library} $\mathcal{L} = \{f_k\}_{k=1}^{M_t}$, where $M_t$ is the number of historical keyframes available before generating shot $s_t$. The library establishes explicit cross-modal mappings to the active registry $\mathcal{R}$. Rather than compressing historical video dynamics into holistic visual embeddings, the agent dynamically appends informative instances to $\mathcal{L}$, forming a high-diversity candidate reference pool for downstream complementary matching.

\paragraph{Library Maintenance Dynamics.} Inspired by the visual caching philosophies in StoryMem~\citep{zhang2025storymem}, we store representative keyframes as reusable references. However, to preserve fine-grained asset diversity for subsequent complementary retrieval, our framework executes a deliberately relaxed caching protocol via a two-stage sequential gate:
\begin{enumerate}
    \item \textbf{Fidelity Filtering:} Candidate frames from the newly synthesized clip are first evaluated via HPSv3~\citep{ma2025hpsv3}. Visually degraded, blurred, or artifact-heavy frames are discarded to ensure reference quality.
    \item \textbf{Temporal Drift Detection:} For the remaining candidate pool, we compute semantic cosine distance using CLIP image features~\citep{radford2021clip}. A qualified frame is committed to $\mathcal{L}$ as a new keyframe $f_k$ if and only if its similarity to the \textit{immediately preceding accepted keyframe within the same shot} falls below a specific threshold, signaling a distinct transition in the visual state.
\end{enumerate}
By comparing frames only with local within-shot history rather than performing aggressive global cross-shot deduplication, this protocol preserves diverse visual perspectives and gives the downstream retriever richer choices.

\paragraph{Dual-Level Structural Annotation.} Once a keyframe $f_k$ is committed to $\mathcal{L}$, a VLM processes it to produce a dual-level semantic representation for downstream use:
\begin{enumerate}
    \item \textbf{Element-Level Annotations ($\mathbf{A}_k$):} The VLM executes closed-set entity verification against $\mathcal{R}$. For each tracked element $e_i \in \mathcal{R}$, $\mathbf{A}_k$ records its presence indicator $a_{k,i} \in \{0, 1\}$, a localized bounding box $b_{k,i}$, and a categorical reference fidelity label $q_{k,i} \in \{\texttt{full}, \texttt{partial}, \texttt{weak}\}$. A \texttt{full} element is complete and clearly identifiable; for a character, it additionally requires a clear frontal face. A \texttt{partial} element is identifiable but occluded, truncated, or, for a character, lacks a clear frontal face, whereas a \texttt{weak} element provides insufficient detail for reliable identity reference. For global environmental backdrops ($\text{type}_i = \text{scene}$), the spatial region $b_{k,i}$ defaults to the full image canvas.
    \item \textbf{Holistic Frame Description ($\mathbf{D}_k$):} Conditioned on the narrative context, the VLM produces a comprehensive textual summary detailing the global composition, artistic style, core action, and plot-level relevance of the entire frame.
\end{enumerate}
This two-tiered design decouples symbolic subset optimization from prompt context alignment. Structurally, the element-level matrix $\mathbf{A}_k$ translates each frame into verifiable evidence for greedy retrieval, explicitly revealing which elements a frame covers and which excluded assets it might introduce. Semantically, the holistic descriptions $\mathbf{D}_k$ provide natural-language context, enabling the prompt agent to articulate \textit{how} the reference frame should be interpreted by the frozen generator.

\subsection{Element-Aware Complementary Retrieval}
\label{subsec:retrieval}
Standard similarity-based Top-$K$ retrieval~\citep{hu2026longliverag,an2025onestory,wei2026memento} frequently selects redundant historical frames depicting identical salient elements, leaving other critical story elements uncovered while compounding historical noise. To circumvent this, we formalize reference selection as a \textit{complementary coverage problem}, seeking a compact subset of keyframes that collectively maximizes target-element coverage while systematically suppressing contextual leakage.

\paragraph{Greedy Coverage Formulation.} For a target shot $s_t$, the semantic query vector $\mathbf{Y}_t$ projects the registry elements into three disjoint constraint sets: required ($R_t = \{e_i \mid y_{i,t}=\texttt{REF}\}$), excluded ($X_t = \{e_i \mid y_{i,t}=\texttt{EXC}\}$), and optional ($O_t = \{e_i \mid y_{i,t}=\texttt{OPT}\}$). Let $\mathcal{F} = \mathcal{L}$ represent the candidate historical keyframes. Leveraging the frame annotations $\mathbf{A}_k$ from the library, we denote the set of visible elements in frame $f$ as $\mathcal{A}(f) = \{e_i \in \mathcal{R} \mid a_{f,i} = 1\}$. 

Our framework greedily constructs a compact reference set $S$ ($|S| \le K$). At each round, Algorithm~\ref{alg:coverage} rescores every unselected candidate according to the dynamically updated set $C$ of already covered required elements. The weights satisfy $w_{\mathrm{uncovered\ REF}} > w_{\mathrm{covered\ REF}} \ge 0$, $w_{\mathrm{OPT}} \ge 0$, and $w_{\mathrm{EXC}} < 0$. Each visible element contributes according to its state: uncovered required elements receive positive rewards, covered or optional elements receive a small or zero reward, and excluded elements incur a negative contribution.

\begin{algorithm}[h]
\caption{Complementary reference retrieval}
\label{alg:coverage}
\begin{algorithmic}[1]
\REQUIRE Candidate frames $\mathcal{F}$; annotated elements $\mathcal{A}(f)$ and fidelity labels $q_{f,e} \in \{\texttt{full},\texttt{partial},\texttt{weak}\}$; required elements $R_t$; excluded elements $X_t$; optional elements $O_t$; the maximum reference count $K$
\ENSURE Selected reference set $S$ and covered required-element set $C$
\STATE $S \leftarrow \emptyset$, $C \leftarrow \emptyset$
\WHILE{$|S| < K$ \text{and} $\mathcal{F}\setminus S \neq \emptyset$}
\FORALL{$f \in \mathcal{F}\setminus S$}
\STATE $\mathrm{score}(f) \leftarrow 0$
\FORALL{$e \in \mathcal{A}(f)$}
\IF{$e \in R_t \setminus C $}
\STATE $\mathrm{score}(f) \leftarrow \mathrm{score}(f) + w_{\mathrm{uncovered\ REF}}$
\ELSIF{$e \in R_t \cap C $}
\STATE $\mathrm{score}(f) \leftarrow \mathrm{score}(f) + w_{\mathrm{covered\ REF}}$
\ELSIF{$e \in X_t$}
\STATE $\mathrm{score}(f) \leftarrow \mathrm{score}(f) + w_{\mathrm{EXC}}$
\ELSIF{$e \in O_t$}
\STATE $\mathrm{score}(f) \leftarrow \mathrm{score}(f) + w_{\mathrm{OPT}}$
\ENDIF
\ENDFOR
\ENDFOR
\STATE $f^\star \leftarrow \arg\max_{f\in \mathcal{F}\setminus S}\mathrm{score}(f)$
\STATE $S \leftarrow S \cup \{f^\star\}$
\STATE $C \leftarrow C \cup \{e \in \mathcal{A}(f^\star)\cap R_t: q_{f^\star, e} = \texttt{full}\}$
\STATE Record score, covered elements, optional elements, and excluded elements for $f^\star$
\ENDWHILE
\RETURN $S, C$
\end{algorithmic}
\end{algorithm}

\paragraph{Fidelity Modulation and Interpretability.} Element-level coverage in $C$ is strictly determined by the VLM-annotated fidelity label $q_{k,i}$. An element $e$ is appended to $C$ only when $q_{k,i}=\texttt{full}$, meaning that it is complete and clearly identifiable; for a character, this additionally requires a clear frontal face. Elements labeled \texttt{partial} or \texttt{weak} can still contribute to a frame's retrieval score, but they are not considered fully covered. This allows subsequent rounds to retrieve alternative frames that depict the same required element more clearly.

Following visual subset selection, an LLM generates natural-language \textit{reference guidance} for each $f_k \in S$. Conditioned on the target shot prompt $p_t$, the shot-level element states $\mathbf{Y}_t$, the frame's element annotations $\mathbf{A}_k$, and its holistic description $\mathbf{D}_k$, this guidance directs the frozen video generator on how to interpret the frame for the current scene. By combining objective element-level instructions, holistic frame descriptions, and contextual guidance, the final prompt ensures fine-grained cross-shot alignment without retraining.

\subsection{Unified Prompt Assembly and Video Generation}
\label{subsec:assembly}
The selected historical evidence, structured element states, and textual instructions are compiled into a unified multimodal prompt to provide explicit, grounded conditioning for the frozen video generator.

\paragraph{Structured Prompt Composition.} To prevent ambiguous interpretation of the attached references, our framework enforces a structured prompt layout consisting of five functional blocks: (i) \textit{current generation task}, (ii) \textit{prefix story context}, (iii) \textit{shot-level element states} ($\mathbf{Y}_t$), (iv) \textit{per-reference grounding instructions}, and (v) \textit{global generation constraints}.

This architecture converts raw reference pixels into explicit, verifiable generation boundaries. Because a historical keyframe can simultaneously depict a required element and an obsolete asset, the prompt explicitly delineates both positive reference guidance and negative suppression boundaries. Element-level instructions specify which characters, objects, and scenes should be preserved or suppressed, while holistic descriptions $\mathbf{D}_k$ and reference guidance provide sequence-level semantic coordination.

\begin{figure}[t]
\centering
\includegraphics[width=\columnwidth]{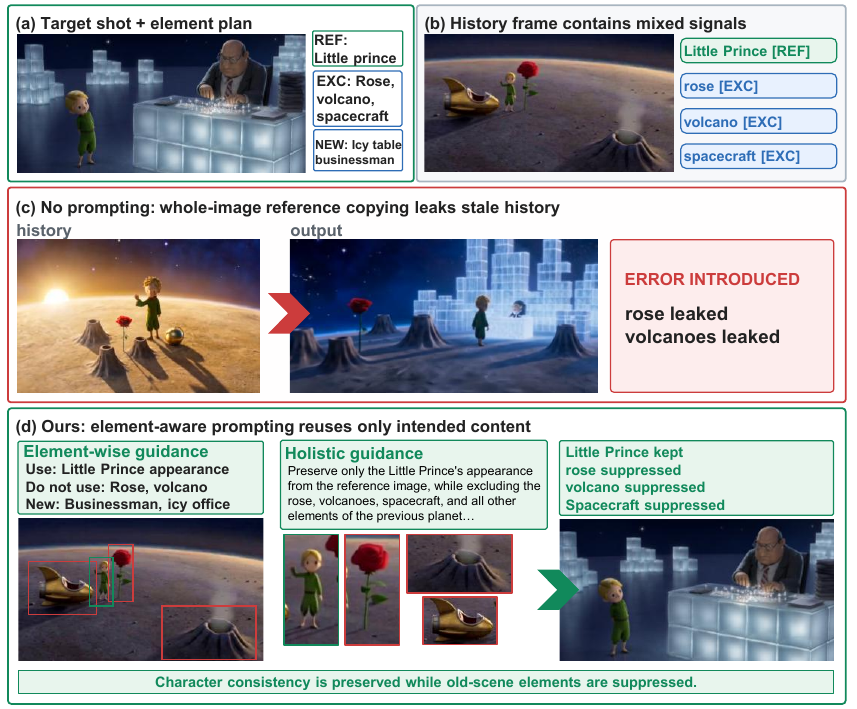}
\caption{Effect of grounded prompting. Without grounding, obsolete scene elements leak into the target shot; ours preserves the character while suppressing conflicting context.}
\label{fig:grounded_prompting_example}
\end{figure}

Figure~\ref{fig:grounded_prompting_example} illustrates this disambiguation. The element plan marks the reusable character as \texttt{REF} and the obsolete scenery as \texttt{EXC}; the prompt agent translates these states into per-reference instructions that preserve character appearance without copying unrelated context.

\paragraph{Video Generation and Sequential Assembly.} The compiled multimodal prompt is sent to the frozen short-video generator. Retrieved references provide long-range visual consistency for both shot types, while the cut indicator $c_t$ determines how local continuity and assembly are handled:
\begin{itemize}
    \item \textbf{Cut Shots ($c_t = \text{true}$):} The generated clip begins a new scene and is directly appended to the preceding sequence.
    \item \textbf{Continuous Shots ($c_t = \text{false}$):} The final segment of $s_{t-1}$ serves as a video prefix to continue the preceding motion and camera trajectory. During assembly, two RIFE-interpolated frames~\citep{huang2022rife} replace the boundary frames of $s_{t-1}$ and $s_t$ to reduce boundary flicker.
\end{itemize}

\subsection{System Properties: Interpretability and Editability}
\label{subsec:properties}
Unlike opaque, end-to-end multi-shot video generation pipelines, our element-aware framework exposes explicit structural reasoning artifacts at every stage, including shot-level element states ($\mathbf{Y}_t$), retrieved complementary reference subsets ($S$), the covered-element set ($C$), grounded multimodal prompts, and keyframe annotations ($\mathbf{A}_k$). This granular transparency directly yields three critical architectural advantages:

\begin{itemize}
    \item \textbf{Attributable Diagnosis:} Generative failures become systematically traceable and diagnosable. For instance, if an obsolete element or a mismatched character identity leaks into a target shot, the user or agent can isolate whether the anomaly stems from agentic element tracking, complementary reference retrieval, or unified prompt assembly.
    \item \textbf{Auto-Reflectivity:} The explicit symbolic boundaries enable autonomous self-reflection and pre-generation verification. The orchestration agent can inspect its reference-coverage matrix, evaluate constraint adherence before invoking the frozen generator, and iteratively refine instructions to maximize target-element fidelity.
    \item \textbf{Human-in-the-Loop Editability:} By modularizing the narrative pipeline into human-interpretable intermediate states, our framework supports manual intervention at runtime. A human director can edit the active element registry, modulate constraint vectors, or replace retrieved references to guide story progression through a controllable and iteratively refinable generation loop.
\end{itemize}

\section{Experiments}
% We study whether retrieval-augmented multimodal prompting improves long-form video generation through two questions:
% \begin{itemize}
% \item \textbf{RQ1:} Does our framework improve cross-shot consistency under a fixed short-video generation backend?
% \item \textbf{RQ2:} How much do visual element planning, complementary coverage retrieval, and grounded reference prompting each contribute?
% \end{itemize}

\subsection{Benchmark and Metrics}

We adopt the \textit{Cross-Shot Consistency} evaluation from EntityBench~\citep{he2026entitybench}, which measures whether recurring characters, objects, and scenes remain consistent across shots. It combines two complementary protocols. First, DINOv2 embedding similarity compares character and object appearances with their per-entity centroids, while a transition-boundary metric measures continuity across scene-internal cuts. Second, LLM pairwise judging compares each non-anchor appearance with a centroid-representative anchor using type-specific criteria, reporting both accuracy and fine-grained similarity; locations are evaluated from full frames with camera-invariant instructions to accommodate viewpoint changes and partial views. The 21 metrics are grouped into \textit{DINOv2 Similarity} (3),
\textit{LLM Characters} (6), \textit{LLM Objects} (6), and
\textit{LLM Scenes} (6), with \textit{Overall} denoting their mean.
EntityBench additionally reports intra-shot metrics including quality and prompt following, which we omit as they
primarily evaluate the base video generator rather than cross-shot
reference construction.

\paragraph{Experimental Settings.}
\textbf{Dataset construction and generation.} EntityBench contains 140 long-range multi-shot episodes. We use an LLM to filter scripts with obvious copyright or sensitive-content risks, yielding 105 eligible episodes, and randomly sample 20 Easy, 10 Mid, and 5 Hard episodes as our test set. Every shot is generated at 480P for 5 seconds. To comply with the moderation constraints of Seedance~2.0, all methods prepend the same instruction requiring non-photorealistic 2D animated human faces and original, non-copyrighted character designs. No predefined character sheets or external reference images are supplied; all visual evidence is generated from the script during inference.

\textbf{Method-specific implementation.} All explicit workflows use their default reference-construction procedures with Seedance~2.0 as the common video backend. ViMax and VideoMemory use Seedream~5.0 to synthesize their dedicated entity references. Training-based baselines use the released pretrained weights and repository-default inference settings on an NVIDIA RTX PRO 6000 GPU. For a fair StoryMem reproduction, we preserve its original keyframe maintenance policy but adapt its 3 Sink + 7 Recent budget to 3 Sink + 6 Recent because Seedance~2.0 accepts at most nine images. Our method retrieves at most five historical frames per shot. Its element weights are 3.0, 2.0, and 1.5 for characters, scenes, and objects; uncovered and covered required elements receive weights 1.0 and 0.2, optional elements 0.1, and excluded elements $-0.1$. The corresponding reference-quality weights for full, partial, and weak evidence are 1.0, 0.2, and 0.1.

\textbf{Keyframe library and evaluation.} StoryMem stores flat keyframe files, retains at most three diverse keyframes per shot under HPSv3 $\geq 3.0$ and CLIP similarity $<0.9$, deduplicates against the complete history, and restricts inference to its Sink--Recent window. In contrast, our library retains up to six diverse keyframes per shot under HPSv3 $\geq 2.5$ and CLIP similarity $<0.95$, only filters near-duplicate adjacent additions, preserves all historical candidates, and attaches VLM element-presence boxes, reference-quality labels, and holistic descriptions. We evaluate all methods using the official EntityBench implementation, replacing its default Gemini judge with Doubao Seed2.1 Turbo uniformly for every method.

\paragraph{Compared Methods.}
We compare against seven representative baselines from recent advances in multi-shot video generation, organized into two groups. The first comprises training-based methods with dedicated generation models: CineTrans~\citep{wu2025cinetrans}, HoloCine~\citep{meng2025holocine}, LongLive2.0~\citep{chen2026longlive2}, and ShotStream~\citep{luo2026shotstream}. The second evaluates explicit reference-construction workflows using the same commercial video API. VideoMemory~\citep{zhou2026videomemory} and ViMax~\citep{huang2026vimax} first synthesize dedicated reference images and then condition video generation on them, whereas StoryMem~\citep{zhang2025storymem} and our approach use keyframes mined from generated history. Although the official StoryMem implementation adapts Wan2.2 with LoRA for multi-reference video generation, its memory construction and keyframe-selection procedures are explicit and deterministic. For a fair comparison, our \textit{StoryMem} reproduction preserves these procedures while replacing the video generator with the same Seedance2.0 backend.

\subsection{Results}

\begin{table}[t]
\centering
{\small
\setlength{\tabcolsep}{3pt}
\begin{tabular}{@{}lrrrrr@{}}
\toprule
Method & Overall & DINO & Char. & Obj. & Scene \\
\midrule
\multicolumn{6}{@{}l}{\textit{Training-based methods}} \\
CineTrans & 0.2964 & 0.5659 & 0.2080 & 0.2235 & 0.3228 \\
HoloCine & 0.4471 & 0.6268 & 0.3125 & 0.5219 & 0.4171 \\
LongLive2.0 & 0.6512 & \textbf{0.8690} & 0.5639 & 0.6464 & 0.6344 \\
ShotStream & 0.6703 & \underline{0.8280} & 0.5789 & 0.5767 & 0.7766 \\
\midrule
\multicolumn{6}{@{}l}{\textit{Explicit workflows (all reproduced with Seedance2.0 backend)}} \\
VideoMemory & 0.7199 & 0.6686 & 0.6593 & 0.7128 & 0.8134 \\
ViMax & 0.7427 & 0.7244 & 0.7864 & 0.7699 & 0.6811 \\
StoryMem & \underline{0.8425} & 0.8078 & \underline{0.8457} & \underline{0.8531} & \underline{0.8461} \\
Ours & \textbf{0.8670} & 0.8249 & \textbf{0.8930} & \textbf{0.8780} & \textbf{0.8510} \\
\bottomrule
\end{tabular}
}
\caption{Cross-Shot Consistency on the same 35-episode EntityBench subset. Overall averages all 21 metrics. Best and second-best results are shown in bold and underlined, respectively.}
\label{tab:cross-shot-main}
\end{table}

As shown in Table~\ref{tab:cross-shot-main}, our method achieves the best \textit{Overall} score of 0.8670 and leads all three LLM-judged categories, with scores of 0.8930 for character consistency, 0.8780 for object consistency, and 0.8510 for scene consistency. These results demonstrate its effectiveness in preserving diverse recurring visual elements across shots.

The results also reveal a metric-dependent pattern. Training-based methods such as LongLive2.0 and ShotStream are particularly competitive in DINOv2 similarity. Their model-internal propagation of history through cross-shot attention, caches, or related memory mechanisms may better preserve embedding-level appearance, and their training objectives may reinforce this effect. However, their lower LLM scores indicate that high feature similarity does not necessarily imply preservation of the correct character attributes, object states, or scene semantics. In contrast, the explicit workflows generally perform better under these fine-grained multimodal judgments, suggesting an advantage from explicitly organizing and grounding historical evidence.

Among methods using the same Seedance2.0 backend, the historical-keyframe approaches, StoryMem and ours, outperform the generated-reference approaches, VideoMemory and ViMax, across all four categories. Dedicated entity images provide clean canonical references but may differ from appearances produced by the video generator and omit useful co-occurrence or scene context. Historical keyframes retain visual evidence from earlier shots, while our annotation, complementary retrieval, and grounded prompting determine what to reuse and suppress.

\subsection{Ablation Study}

\begin{table}[t]
\centering
{\small
\setlength{\tabcolsep}{3pt}
\begin{tabular}{@{}lrrrrr@{}}
\toprule
Method & Overall & DINO & Char. & Obj. & Scene \\
\midrule
Full & \textbf{0.8670} & \textbf{0.8249} & \textbf{0.8930} & \textbf{0.8780} & \textbf{0.8510} \\
w/o Cover & 0.8482 & 0.7855 & \underline{0.8658} & 0.8623 & 0.8479 \\
w/o Prompting & \underline{0.8491} & 0.8134 & 0.8513 & \underline{0.8741} & 0.8396 \\
CLIP Retrieval & 0.8456 & \underline{0.8233} & 0.8407 & 0.8584 & \underline{0.8488} \\
No Reference & 0.5456 & 0.7241 & 0.4058 & 0.5788 & 0.5628 \\
\bottomrule
\end{tabular}
}
\caption{Cross-Shot Consistency of ablations and retrieval baselines on the same 35-episode subset.}
\label{tab:cross-shot-ablation}
\end{table}

All variants use the same 35 episodes, generation backend, VLM judge, and animation constraints; variants using references also share the same reference budget of 5 images. \textit{Full} denotes the complete pipeline. \textit{w/o Cover} retains visual element planning and grounded prompting but replaces iterative complementary selection with a single-pass Top-$K$ selection based on the same static element-aware frame scores. \textit{w/o Prompting} retains visual element planning and greedy coverage but conditions the generator only on the original shot prompt and selected images, omitting structured element plans and reference-specific guidance. \textit{CLIP Retrieval} keeps the keyframe library and retrieves historical frames using CLIP text--image similarity between the current shot prompt and each candidate frame, without visual element planning, complementary coverage, or reference grounding. \textit{No Reference} uses only the shot prompt and removes historical visual references entirely.

Table~\ref{tab:cross-shot-ablation} demonstrates that \textit{Full} yields the highest \textit{Overall} score and ranks first across all four evaluation criteria. \textit{w/o Cover} lowers \textit{Overall} by 0.0188 and degrades every reported metric, confirming that independently high-scoring frames do not necessarily form a complementary reference set. \textit{w/o Prompting} lowers \textit{Overall} by 0.0179 and character consistency by 0.0417, indicating that selected images remain ambiguous unless the generator is told which evidence to preserve and which historical content to suppress. \textit{CLIP Retrieval} trails \textit{Full} by 0.0214 in \textit{Overall} and 0.0523 in character consistency. Because it jointly removes element planning, element-aware coverage, and reference grounding, this comparison measures the benefit of the integrated element-aware workflow rather than any single component. Finally, \textit{No Reference} reduces \textit{Overall} to 0.5456, demonstrating that text-only prompting is insufficient to preserve recurring entities across independently generated shots.

\subsection{Qualitative Analysis}

Figure~\ref{fig:complementary_example} shows that greedy coverage avoids redundant Top-$K$ frames by favoring uncovered elements, yielding a complementary set that covers all required content. Figure~\ref{fig:grounded_prompting_example} demonstrates how grounding disambiguates a selected frame: without it, the volcanoes and the rose from the Little Prince's former planet leak into his visit to a merchant on a new planet. Our method preserves the Little Prince's appearance while explicitly excluding these obsolete scene elements.

\begin{figure*}[t]
    \centering
    \includegraphics[width=\textwidth]{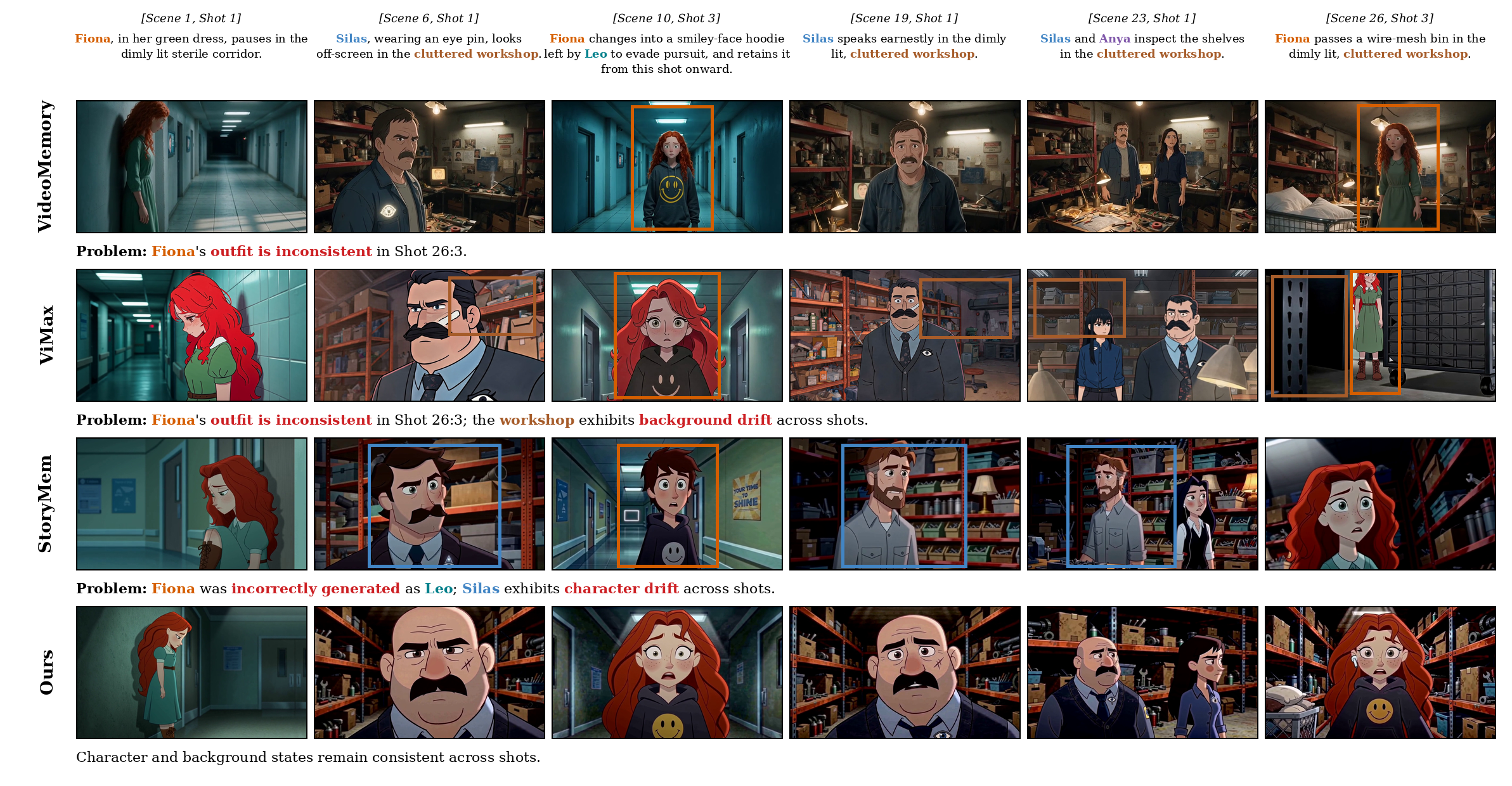}
    \caption{Qualitative comparison of explicit workflows on EntityBench \textit{Hard\_5}; all methods use Seedance~2.0. Colored text and boxes identify the relevant characters and scene. After Fiona changes into Leo's smiley-face hoodie in Scene~10, Shot~3, VideoMemory and ViMax exhibit outfit inconsistency in Scene~26, Shot~3; ViMax also exhibits background drift. StoryMem exhibits character confusion for Fiona and character drift for Silas. Our method preserves the marked character, outfit, and background states.}
    \label{fig:workflow-qual}
\end{figure*}

Figure~\ref{fig:workflow-qual} further compares explicit reference-management workflows under the same Seedance~2.0 backend on the challenging \textit{Hard\_5} episode. Fiona changes from her green dress into Leo's smiley-face hoodie in Scene~10, Shot~3 and should retain this outfit thereafter. VideoMemory and ViMax exhibit outfit inconsistency in the later Fiona shot; ViMax also exhibits background drift in the cluttered workshop. StoryMem produces character confusion by rendering Fiona as Leo, followed by character drift for Silas. In contrast, our framework preserves the intended character, outfit, and scene states across the long-range sequence.

\begin{figure*}[t]
    \centering
    \includegraphics[width=\textwidth]{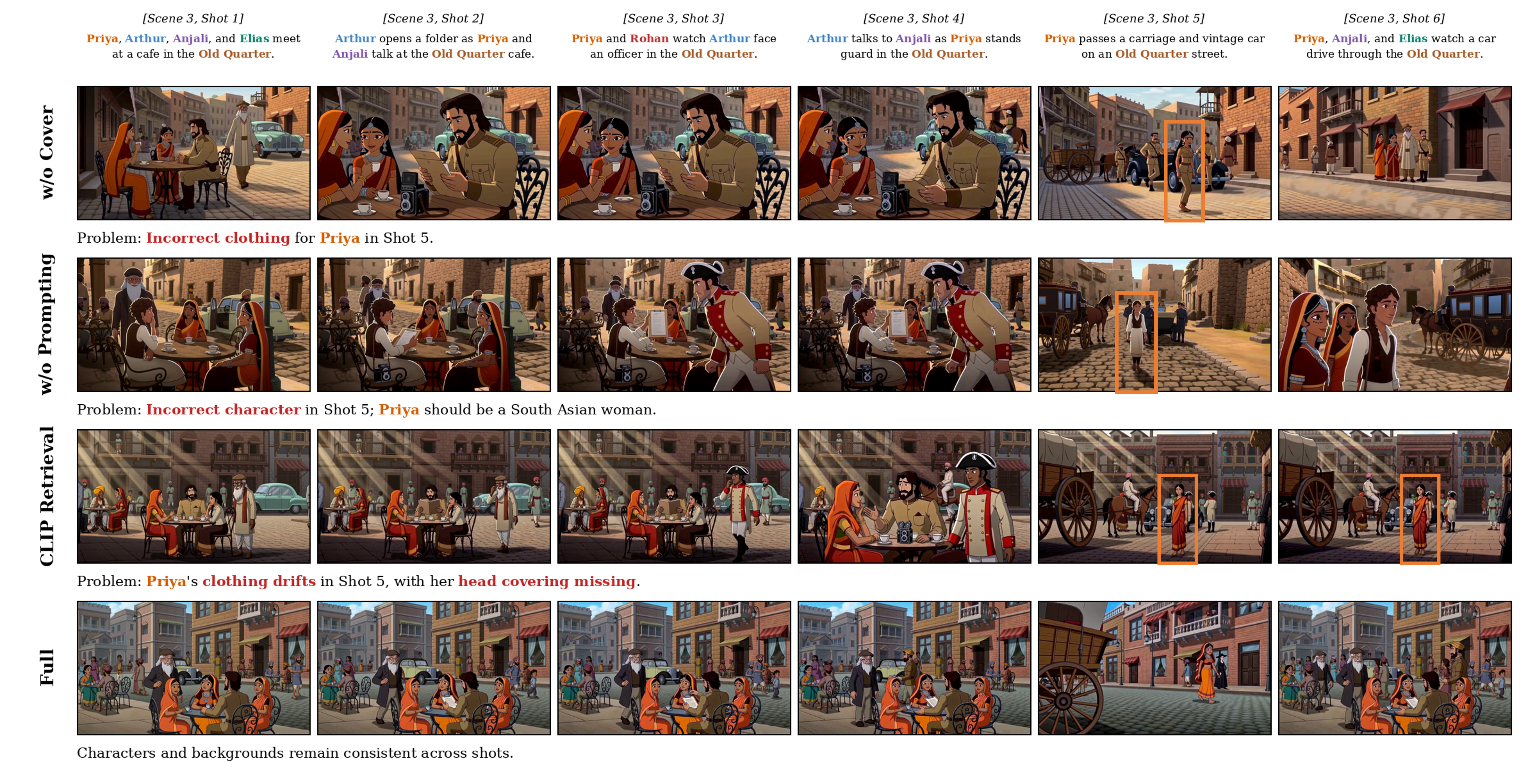}
    \caption{Qualitative ablation on Scene~3 of EntityBench \textit{Easy\_2}. Without complementary coverage, Priya has an outfit inconsistency in Shot~5. Without structured prompting, Priya is incorrectly generated in Shot~5. CLIP retrieval causes clothing drift and loses her head covering. The complete pipeline maintains the intended character and Old Quarter scene states.}
\label{fig:ablation-qual}
\end{figure*}

Figure~\ref{fig:ablation-qual} visualizes the component effects on Scene~3 of \textit{Easy\_2}. Removing complementary coverage selection yields an outfit inconsistency for Priya, while removing structured prompting produces character confusion. Replacing element-aware retrieval with generic CLIP retrieval causes clothing drift and removes Priya's head covering. The complete pipeline maintains both the recurring characters and the Old Quarter scene across the displayed shots.

\section{Conclusion}
We presented Complementary Retrieval-Augmented Prompting, a training-free framework for consistent long-form video generation with frozen short-video generators. Instead of compressing visual history into implicit memory or relying on manually curated references, our framework maintains a text-grounded visual element registry together with a VLM-annotated keyframe library, turning generated history into structured and queryable evidence. Coverage-aware retrieval selects a compact set of complementary frames, while grounded multimodal prompting specifies what each frame should contribute and which stale content should be suppressed. EntityBench experiments demonstrate superior cross-shot consistency, while ablations validate complementary retrieval and reference grounding. Beyond accuracy, the explicit intermediate representations make reference decisions interpretable and editable, supporting interactive creation without a complete script. Separating evidence maintenance, retrieval, and generation also facilitates backend portability and allows failures to be traced to individual pipeline stages rather than hidden within model states. Current limitations include errors in element planning and visual annotation, imperfect prompt adherence by the underlying generator, and constraints imposed by commercial APIs. Future work will improve agentic verification, scale retrieval to longer narratives, and strengthen transfer across video-generation backends and interactive production settings.

% Avoid \flushbottom stretching the flexible bibliography list spacing.
{\raggedbottom
\bibliography{reference}
}


\end{document}